\documentclass[11pt]{article}
\usepackage{eacl2017}
\usepackage{times}
\usepackage{latexsym}
\usepackage{url}
\usepackage{graphicx}
\usepackage{subcaption}
\usepackage{url}
\usepackage{wrapfig}
\usepackage{floatflt}

\usepackage{enumitem}
\usepackage{booktabs}
\usepackage{amssymb}
\usepackage{array}
\usepackage{caption}

\eaclfinalcopy % Uncomment this line for the final submission
\title{Summarizing Dialogic Arguments from Social Media}

\author{Amita Misra, Shereen Oraby, Shubhangi Tandon, Sharath TS, \\
            {\bf Pranav Anand  and  Marilyn Walker}\\
	    UC Santa Cruz\\
	Natural Language and Dialogue Systems Lab \\
    1156 N. High. SOE-3\\
	    Santa Cruz, California, 95064, USA\\
	    {\tt amisra2|soraby|shtandon|sturuvek|panand|mawalker@ucsc.edu} \date{}
}
\begin{document}
\maketitle

\begin{abstract}
Online argumentative dialog is a rich source of information on
popular beliefs and opinions that could be useful to companies as well
as governmental or public policy agencies.  Compact, easy to read,
summaries of these dialogues would thus be highly valuable.  A
priori, it is not even clear what form such a summary should take.
Previous work on summarization has primarily focused on summarizing
written texts, where the notion of an abstract of the text is well
defined. We collect gold standard training data consisting of five human
summaries for each of 161 dialogues on the topics of {\it Gay Marriage},
{\it Gun Control} and {\it Abortion}.  We present several different
computational models aimed at identifying segments of the dialogues
whose content should be used for the summary, using linguistic features and Word2vec features 
with both SVMs and
Bidirectional LSTMs. We show that we can identify the most important
arguments by using the dialog context with a best F-measure of 0.74 for
gun control, 0.71 for gay marriage, and 0.67 for abortion.

\end{abstract}

\section{Introduction}
\label{intro-sec}

Online argumentative dialog is a rich source of information on
popular beliefs and opinions that could be useful to companies as well
as governmental or public policy agencies.  Compact, easy to read,
summaries of these dialogues would thus be highly valuable. 
However, previous work on summarization has primarily focused on summarizing
written texts, where the notion of an abstract of the text is well
defined. 

Work on dialog summarization is in its infancy.  Early work was domain
specific, for example focusing on extracting actions items from
meetings \cite{Murray08summarizingspoken}. \newcite{GurevychStrube04}
applied semantic similarity to Switchboard dialog, showing
improvements over several baseline summarizers. Work on argument
summarization has to date focused on monologic
data. \newcite{Ranade2013} summarize online debates using topic and
sentiment rich features, but their unit of summary is a single debate
post, rather than an extended conversation. \newcite{Wang2016Neural}
generate abstractive one sentence summaries for opinionated arguments
from debate websites using an attention-based neural network model,
but the inputs are well-structured arguments and a central claim
constructed by the editors, rather than user-generated conversations.

\begin{figure}[h!t]
%\begin{scriptsize}
\begin{small}
%{\begin{tabular}{|p{.25in}p{2.4in}|p{.39in}|}

{\begin{tabular}{|p{.27in}p{2.43in}|}
\hline {\bf PostID} & {\bf Turn}% & {\bf Stance} 
\\ \hline \hline

S1-1: & Gays..you wont let me have everything I want so you must hate me. Spoil child..you wont let me have everything I want so you must hate me. \\
%& AGAINST \\ %\hline

S2-1: & And who made you master daddy that you think it is your place to grant or disallow anything to your fellow citizens?\\
% & FOR  \\ %\hline 

S1-2: & Did I say that I was and it is?\\
%& AGAINST \\ %\hline 

S2-2: & You implied it when you compared gays (and their supporters) fighting for rights to spoiled children. For the analogy to work there has to be a parent figure for the gays as well.\\
%& FOR \\ %\hline 
S1-3: & The public is the 'parent' figure and the law makers are ( or should be) the public's servant .\\
%& AGAINST \\ %\hline

S2-3: & This then implies that homosexuals are are not part of the public and the law-makers are not their servants as well, and that you do indeed believe it is your right to allow and disallow things to your fellow citizens. That they are lesser group than you. You just proved your hate. \\
%& FOR \\ %\hline
S1-4: & Homosexuals are a deviant minority. 
%& AGAINST
\\ \hline
\end{tabular}}
\end{small}
%\vspace{-0.2in}
\caption{\label{Gay-Rights-dialog1} Gay Rights Argument.}
%\vspace{-.3in}
\end{figure}

\begin{table*}[t!h]
%\captionsetup{font=scriptsize}
\begin{center}
\begin{small}
\begin{tabular} { |p{10cm} | >{\centering}p{3cm}  |p{.6cm}|}  \hline
\bf Summary Contributors& \bf  {Human Label from Pyramid Annotations} & \bf {Tier Rank}  \\ \hline
 \begin{itemize}[noitemsep,topsep=0pt,parsep=0pt,partopsep=0pt,leftmargin=*]
 \item  S1 says that no one can prove that gun owners are safer than non gun owners.
 \item S1 says no one has been able to prove gun owners are safer than non-gun owners.
  \item S1 points out there is no empirical data suggesting that gun owners are safer than non-gun owners.
  \item S1 states there are no statistics proving owning a gun makes people safer.
 \item  S1 believes that there is no proof that gun owners are safer than non-gun owners.
   \end{itemize}
   & Nobody has been able to prove that gun owners are safer than non-gun owners. &5 \\ \hline
    \begin{itemize}[noitemsep,topsep=0pt,parsep=0pt,partopsep=0pt,leftmargin=*]
 \item They say that if S2 had a family member die from gun violence it might be more significant to them,
 \item  He says if S1 had a personal or family encounter with gun violence, he would feel differently.
 \item  that people who have had relatives die from gun violence have a different attitude.
 \end{itemize} &Family encounters with gun violence changes significance.&3 \\ \hline
 \begin{itemize}[noitemsep,topsep=0pt,parsep=0pt,partopsep=0pt,leftmargin=*]
\item Pro-gun perspective is: on 9/11, 3000 people died without the ability to defend themselves. \end{itemize}& On 9/11, 3000 people died without the ability to defend themselves. &1 \\ \hline

\end{tabular}
\vspace{-.1in}
\end{small}
\caption{\label{contrib-label} Example summary contributors, pyramid labels and tier rank in gun control dialogs}
\end{center}
\vspace{-.2in}
\end{table*}

To our knowledge there is no prior work on summarizing important
arguments from noisy, argumentative, dialogs in online debate such
as that in Figure~\ref{Gay-Rights-dialog1}.  A priori, it is not
even clear what form such a summary should take.  The two conversants
in Figure~\ref{Gay-Rights-dialog1} obviously do not agree: should a
summary give preference to one person's views? Should a summary
be based on decisions about which argument is higher quality, well
structured, more logical, or which better follows theories of
argumentation?

Fortunately, summarization is something that any native speaker can do
without formal training. Thus our gold standard training data consists
of 5 human summaries for each dialog from a corpus of dialogs
discussing {\it Gay Marriage}, {\it Gun Control} and {\it Abortion}.
Arguments that are important to extract to form the basis of summary
content are defined to be those that appear in a majority of human
summaries, as per the Pyramid model \cite{pyramid04}.  We then
aim to learn how to automatically extract these important
arguments from the original dialogs.

We first define several baselines using off-the-shelf summarizers such
as LexRank and SumBasic
\cite{erkan2004lexrank,nenkova2005sumbasic}. Our experiments explore
the effectiveness of combining traditional linguistic features with
Word2Vec in both SVMs and Bidirectional LSTMs. We show that applying
coreference, and representing the context improves
performance. Performance is overall better for the Bidirectional LSTM,
but both models perform better when linguistic features and
argumentative features are combined with word embeddings. We achieve a
best F-measure of 0.74 for gun control, 0.71 for gay marriage, and
0.67 for abortion. We discuss related work in more detail in
Section~\ref{rel-sec} when we can compare it with our approach.

\section{Experimental Method}
\label{sec-ml-results}

\subsection{Data}
\label{corpus-sec}

Our corpus of dialogs and summaries focus on the topics {\it Gay
  Marriage}, {\it Gun Control} and {\it Abortion} from the the
publicly available Internet Argument Corpus (IAC)
\cite{Abbottetal16}. We used the portion of the IAC containing posts
from {\small \url{ http://4forums.com}}.  We use the debate forum
metadata to extract dialog exchanges between pairs of authors with at
least 3 turns per author, in order to represent 2 different
perspectives on an issue. To get richer and more diverse data per
topic containing multiple argumentative claims and propositions, we
ensure that the corpus does not contain more than one dialog per topic
between any particular pair of authors.  The dataset contains 61 gay
rights dialogues, 50 gun control dialogues and 50 abortion
dialogues. 

\begin{table*}
{\begin{tabular}{|p{6.1in}|} \hline
	\small
In this task, you will carefully read part of a dialog where two people are discussing the issue of gun control. Several previous workers have each summarized this dialog, and we have related those summaries by grouping together parts of their summaries that roughly describe the same actions in the dialogue. In this task, you will link these action description groups to sentences in the dialogue.
Each dialog is automatically divided into sentences. Your job is to provide the best action description group for each sentence.

The action description groups are sets of sentences from several summaries that essentially describe the same action in the dialog in different words. Each group has a unique label and you will select the label that best approximates what is happening in the sentence and select a label using the radio button provided with each sentence. 

Please especially note:
\begin{itemize}[noitemsep,topsep=0pt,parsep=0pt,partopsep=0pt,leftmargin=*]
\item More than one sentence can map to same group. For example, two people may say virtually the same thing multiple times. 
\item Not all sentences will have a good group, so if you cannot find any similar set for a sentence, then select None of the labels match in the radio button option.
\item You are expected to read and comprehend the sentence. Since these come from summaries, the action summaries may use very different words from those used in the dialogs. 
\end{itemize}
\\  \hline

\end{tabular}}
\caption{\label{hit-dirs} Directions for Step 3 ({\bf S3} annotation, mapping pyramid labels to sentences.}
\end{table*}

We adopt a three step process to identify useful sentences for
extraction that we briefly summarize here. 

\begin{itemize}
\item {\bf S1:} Dialogs are 
read and summarized by 5 pre-qualified workers on Mechanical
Turk. Since the dialogs vary in length and content we applied a
limit that dialogs with a word count less than 750, must be summarized
by the annotators in 125 words and dialogs with word count greater
than 750 words should be summarized in 175 words. 
\item {\bf S2:} We train undergraduate linguists to use 
the Pyramid method \cite{pyramid04} to identify
important arguments in the dialog; they then construct
pyramids for each set of five summaries. Repeated elements of the five
summaries end up on higher tiers of the pyramid, and indicate the most
important content, as shown in Table~\ref{contrib-label}. This results
in a ranking of the most important arguments (abstract objects)
in a dialog, but the
linguistic representation of these arguments is based on 
the language used in the summaries themselves. 
\item {\bf S3:} To identify the
spans of text in the dialog itself that correspond to the
important arguments, we must map the ranked labels
from the summaries back onto the dialog text.  We recruited
2 graduate students and 2 undergraduates to label each sentence of the
dialog with the best set of human labels from the pyramids.
Table~\ref{hit-dirs} shows the directions for this task.
\end{itemize}

%% \begin{figure}[h]
%% \captionsetup{font=small}
%% \centering
%% \includegraphics[width=2.7in]{alltopic_histogm2.png}
%% \vspace{-.1in}
%% \caption{\label{label-dist-histogm} Frequency of average sentence tier rank across the three topics based on annotator label selection. ``None" annotations are assigned a score of 2.}
%% %\end{wrapfigure}
%% \vspace{-.1in}
%% \end{figure}

We now have one or more labels for each sentence in a dialog, but we
are primarily interested in the {\bf tier rank} of the sentences.  We
group labels by tier and compute the average tier label per sentence.
We define any sentence with an average tier score of 3 or higher as
\textbf{important}.  Thus, steps {\bf S1}, {\bf S2} and {\bf S3} above
are simply carried out to arrive at a well-motivated and theoretically
grounded definition of \textbf{important} argument, and the task we
address in this paper is binary classification applied to dialogs to
select sentences that are important.
%Figure~\ref{label-dist-histogm} shows the
%reduction of data defined by this selection process and 
Table~\ref{tab-corpus-statistics} shows the resulting number of
important sentences for each topic. 
The average Cohen's kappa between the annotators
is respectable, with  a kappa value of 0.68
for gun control, 0.63 for abortion, and 0.62 for 
gay marriage.

\begin{table}[!htb]
\begin{center}
%\begin{scriptsize}
\begin{small}
\begin{tabular}{|p{1.9cm}||p{1.6cm}| p{1.9cm} | }
\hline
Topic & Important  & Not Important   \\ \hline
Gun Control & 1010 & 1041 \\ \hline
Gay Marriage & 1311    & 1195    \\ \hline
Abortion &   849   & 1203   \\ \hline
\end{tabular}
%%\vspace{-.15in}
\caption{\label{tab-corpus-statistics}
Sentence distribution in each domain. }
\end{small}
%\end{scriptsize}
\end{center}
\vspace{-.2in}
\end{table}

\subsection{Baselines}

We use several off-the-shelf extractive summarization engines
(frequency, probability distribution and graph based) from the python package sumy 
\footnote{{\small {\tt https://pypi.python.org/pypi/sumy}}} to provide
a baseline for comparison with our models.  To enable direct
comparison, we define a sentence as \textbf{important} if it appears
in the top $n$ sentences in the output of the baseline summarizer,
where $n$ is the number of \textbf{important} sentences for the
dialog as defined by our method.
 
\noindent{\bf SumBasic.} \newcite{nenkova2005sumbasic} show that
content units and words that are repeated often are 
likely be mentioned in a human summary, and that frequency is a
powerful predictor of human choices in content selection for
summarization. SumBasic uses a greedy search approximation with a
frequency-based sentence selection component, and a component to
re-weight the word probabilities in order to minimize
redundancy. 

\noindent{\bf KL divergence Summary.} This approach is based on
finding a set of summary sentences which closely match the document
set unigram distribution. It greedily adds a sentence to a summary as
long as it decreases the KL Divergence  \cite{Haghighi2009}.

\noindent{\bf  LexRank.} This method is a degree-based method of computing
centrality that is used for extractive summarization and has shown to
outperform centroid-based methods on DUC evaluation tasks.  It
computes sentence importance based on eigenvector centrality in a
graph where cosine similarity is used for sentence adjacency weights
in the graph \cite{erkan2004lexrank}.

\begin{figure}[th!]
%\begin{scriptsize}
\begin{small}
{\begin{tabular}{|p{2.8in}|}

%{\begin{tabular}{ |p{.27in}| p{2.0in}| p{2.2.cm}|}
\hline {\bf  Summary Sentences selected by human annotators }  \\ \hline \hline
Nobody has been able to prove that gun owners are safer than non-gun owners.  \\  \hline
You can play around with numbers to make the problem seem insignificant.  \\  \hline
I suppose you could also say that only 3,000 people died in 9/11 and use your logic to say that it 's only a small problem.  \\  \hline
Perhaps if somebody in your family had died of gun violence you would have a different attitude. \\  \hline
Nobody has been able to prove that non-gun owners are safer than gun owners.  \\  \hline
So if you can not prove things one way or the other why try to infringe on my rights?  \\  \hline
I did n't say that it ca n't be proven one way or the other.  \\  \hline
I just said you ca n't prove that gun owners are safer.  \\  \hline
Using illogic , skewed statistics , revisionist history all in an attempt to violate my constitutional rights , that would be you and other gun grabbers who are trying to infringe on law abiding citizens rights.  \\  \hline
Show me in the Constitution where it says that making an illogical argument is a violation of somebody 's rights.  \\  \hline
You and your ilk are doing everything in your power to implement your '' victim disamament '' program in '' violation '' of my civil rights.  \\  \hline
No different than '' jim crow '' laws and other unconstitutional drivel. \\  \hline

\end{tabular}}
\end{small}
%\vspace{-0.2in}
\caption{\label{gold-std} Human selected summary sentences for a gun control dialogue.}
%\vspace{-.3in}
\end{figure}

\begin{figure}[th!]
%\begin{scriptsize}
\begin{small}
{\begin{tabular}{|p{2.8in}|}

%{\begin{tabular}{ |p{.27in}| p{2.0in}| p{2.2.cm}|}
\hline  {\bf  Summary sentences selected by LexRank}  \\ \hline 

Show me in the Constitution where it says that making an illogical argument is a violation of somebody 's rights.  \\  \hline
Nobody has been able to prove that gun owners are safer than non-gun owners. \\  \hline
I just said you ca n't prove that gun owners are safer. \\  \hline
{\bf Wow that is easy}. \\  \hline
{\bf At least have the courage to say it ... .} \\  \hline
{\bf Witch hunt. }\\  \hline
No different than '' jim crow '' laws and other unconstitutional drivel. \\  \hline
So if you can not prove things one way or the other why try to infringe on my rights? \\  \hline
{\bf Oh, stop your witch hunt. }\\  \hline
You can play around with numbers to make the problem seem insignificant. \\  \hline
Using illogic, skewed statistics, revisionist history all in an attempt to violate my constitutional rights, that would be you and other gun grabbers who are trying to infringe on law abiding citizens rights. \\  \hline
I suppose you could also say that only 3,000 people died in 9/11 and use your logic to say that it 's only a small problem. \\  \hline

\end{tabular}}
\end{small}
%\vspace{-0.2in}
\caption{\label{lex-summ} Lex Rank selected sentences for a gun control dialogue.}
%\vspace{-.3in}
\end{figure}

Figures~\ref{gold-std} and \ref{lex-summ} show our gold standard
summary and the summary sentences selected by LexRank for the same
dialog. LexRank identifies many of the important sentences, but it
also includes a number of sentences which cannot be used to construct
a summary such as {\it "Wow that is easy"}. The baseline outputs in
general suggest that frequency or graph similarity alone leave room
for improvement when predicting important sentences in user-generated
argumentative dialogue.

\subsection{Features}

Most formal models of argumentation have focused on carefully crafted
debates or face-to-face exchanges. However, as the `bottom-up'
argumentative dialogs in online social networks are far less logical
\cite{gabbriellini2013ms,Toni2011bottom}, and the serendipity of the
interactions yields less rule-governed conversational turns, ones that
violate even the rules of naturalistically grounded argument models
\cite{WaltKrab95}. This makes it difficult to construct useful
theoretically-grounded features. In place of that enterprise, we
exploit more conventional summarization, sentiment, word class, and
sentence complexity features.

We also construct features sensitive to dialogic context. The
theoretical literature discusses the ways in which dialogic
argumentation shows different speech act uses than in less
argumentative genres \cite{Budzynska2011speechact,Jacobs1992},
including the fact that arguments in these conversations are
frequently smuggled in via non-assertive speech acts (e.g., hostile
questions). Inspired by this, we implement three basic methods for
dialogic context: we extract the dialog act tag and some word class
class information from the previous sentence; we extract a
rough-grained measure of a sentence's position within a turn; and we
use coreference chains to resolve anaphora in a sentence to acquire a
(hopefully) more contentful antecedent. Below, we describe these
features in more detail.

\noindent{\bf Google Word2Vec}: Word embeddings from word2vec~\cite{Mikolovetal13} are popular for expressing semantic
relationships between words. Previous work on argument mining has developed methods using word2vec that are effective for argument recognition \cite{Habernalemnlp2015}. We created a 300-dimensional vector by filtering stopwords and punctuation and then averaging the word embeddings from Google's word2vec model for the remaining words.

\noindent{\bf GloVe Embeddings:} GloVe is an unsupervised algorithm
for obtaining vector representations for words
\cite{pennington2014glove}.  These pre-trained word embeddings are 100
dimensional vectors and each sentence is represented as a
concatenation of word vectors. We use GloVe embeddings to initialize
our Long Short-Term Memory (LSTM) models as glove embeddings have been
trained on web data, and in some cases work better than Word2Vec
\cite{stojanovski2016finki}.

\noindent{\bf Readability Grades:} We hypothesized that contentful
sentences were more likely to be complex. To measure that, we used
readability grades, which calculate a series of linear regression
measures based on the number of words, syllables, and sentences. We
used 7 readability measures\footnote{{\small {\tt
      https://pypi.python.org/pypi/readability}}} Flesch-Kincaid
readability score, Automated Readability Index, Coleman-Liau Index,
SMOG Index, Gunning Fog index, Flesch Reading Ease, LIX and RIX.

\noindent{\bf LIWC:} The Linguistics Inquiry Word Count (LIWC) tool has been
useful in previous work on stance detenction~\cite{PennebakerFrancis01,SomasundaranWiebe09,HasanNg13}, and we suspected it would help to distinguish personal conversation from substantive analysis.  It classifies words into different categories based on thought processes, emotional states, intentions, and motivations. For each LIWC category, we computed an aggregate frequency score for a sentence. Using these categories we aim to capture both the style and the content types in the argument. Style words are linked to measures of people's social and psychological worlds while content words are generally nouns, and regular verbs that convey the content of a communication. To capture additional contextual information, we computed the LIWC score of the previous sentence.

\noindent{\bf Sentiment:} Sentiment features have shown to be useful
for argumentative claim identification, and here too we suspected that
name-calling and the like could be flagged by sentiment features. We
used the Stanford sentiment analyzer from \cite{socher2013EMNLP} to
compute five sentiment categories (very negative to very positive) per
sentence.

\noindent{\bf Dialog Act of Previous Sentence (DAC):} We hypothesized
that \textbf{important} sentences may be more likely in response to
particular dialog acts, like questions, e.g. a question may be followed
by an explanation or an answer. To identify if a previous sentence was
a question, we combined the tags into two categories indicating
whether the previous sentence was a question type or not. We
implemented a binary PreviousSentAct feature which used Dialog Act
Classification from NLTK \cite{LoperBird02}.

\noindent{\bf Sentence position:} We divide a turn into thirds and
create an integral feature based on which third a sentence is located
in the turn.

\noindent{\bf Coref}: In the hope that coreference resolution would
help ground utterance semantics, we replaced anaphoric words with
their most representative mention obtained using Stanford coreference
chain resolution \cite{manning-EtAl:2014:P14-5}.

\subsection{Machine Learning Models}
\label{sec-ml-results}
We reserved 13 random dialogs in each topic for our test set, using
the rest as training. Sentences were automatically split. This led to
several sentences consisting essentially of punctuation, which were
removed (filter for sentences without a verb and at least 3 dictionary
words.) For learning, we created a balanced training and test set by
randomly selecting an equal number of sentences for each class, giving
the following combinations: 1236 train and 462 test sentences for
abortion, 1578 training and 534 test for gay marriage and 1352
training and 476 test for gun control. We 
use two machine learning models. 

\noindent{\bf SVM.}  We use Support Vector Machines with a linear
Kernel from Scikit-learn \cite{Pedregosaetal11} with our theoretically
motivated linguistic features and uses cross validation for parameter
tuning and the second is a combination Bidirectional
LSTM. 

\noindent{\bf CNN + BiLSTM.}  A combination of Convolutional and
Recurrent Neural Networks has been used for sentence representations
\cite{Wang2016CombinationOC} where CNN is able to learn the local
features from words or phrases in the text and the RNN learns
long-term dependencies. Using this as a motivation, we include a
convolutional layer and max pooling layer before the input is fed into
an RNN. The model used for binary classification consists of a 1D
convolution layer of size 3 and 32 different filters. The convolution
layer takes as input the GloVe embeddings. A bidirectional LSTM layer
is stacked on the convolutions layer and then concatenated with
another layer of bidirectional LSTM: different versions are used with
different features and feature combinations as shown in
Table~\ref{results-tab} and described further below. The outputs of
the LSTM are fed through a sigmoid layer for binary
classification. LSTM creates a validation set by a 4 to 1 random
selection on the training set. Regularization is performed by using a
drop-out rate of 0.2 in the drop-out layer.  The model is optimized
using the Adam \cite{kingma2014adam} optimizer. The deep network was
implemented using the Keras package \cite{chollet2015keras}.

\begin{table*}[ht!]
%\captionsetup{font=scriptsize}
\begin{small}
\begin{tabular}{|p{0.18cm} | p{1.3cm} | p{2.7cm} | p{1.2cm}    |   p{1.2cm} || p{1.2cm}    |   p{1.2cm} || p{1.2cm}    |   p{1.2cm}  | }  \hline

      & & &\multicolumn{2}{c||}{\bf Gun Control} & \multicolumn{2}{c||} {\bf Gay Marriage} & \multicolumn{2}{c|}{\bf Abortion} \\  \hline
 {\bf ID}&Classifier & {\bf Features}  & {\bf F-weight Avg.} & {\bf F-weight Avg. Coref} & {\bf F-weight Avg. } &  {\bf F-weight Avg. Coref} &  {\bf F-weight Avg.} &  {\bf F-weight Avg. Coref }\\  \hline

1A &Baseline& KL-SUM ({\bf KL ) }& 0.51 & &  0.52    &  & 0.47  & \\  \hline
1B &Baseline&SumBasic ({\bf SB ) }&0.53 & & 0.57   &   & 0.49 & \\  \hline
1C & Baseline& Lex-Rank ({\bf LR ) }&0.58 & &  0.58    & &0.59  & \\  \hline
\hline
2 &SVM& Dialog Act ({\bf DAC)} & 0.61 & 0.60 &0.58&0.58&0.42&0.41\\  \hline
3 & SVM&Word2Vec                                           & 0.65 & 0.65&0.63 &0.56 &0.58  & 0.58\\  \hline
4 & SVM &Readability {\bf (R)}                            & 0.64 & 0.67 &0.68 &0.68&0.63& 0.64\\  \hline
5 &SVM& LIWC current sentence {\bf (LC) }     & 0.72 & {\bf 0.74} & {\bf 0.69} &0.66& 0.64 & 0.63 \\  \hline
6 &SVM& Sentiment {\bf (SNT) }                       & 0.66 &  &0.62 && 0.61 &  \\  \hline
7 &SVM& Sentence Turn {\bf (ST) }                   & 0.61 & 0.61 &0.40 &0.40& 0.33& 0.33 \\  \hline
8 & Bi LSTM  &                                                   & 0.68 & 0.69&0.63&0.58 &0.64& {\bf 0.65} \\  \hline
\hline

 \multicolumn{9} {c }{\bf Feature Combinations }  \\  \hline \hline

9 & SVM &LIWC current + previous {\bf (LCP) } &  0.73& 0.72 & 0.66 &0.67&  0.61 & 0.61\\  \hline
10 & SVM &LCP  +  R                & 0.73 &     0.73        &0.70 & 0.68 &  0.61 & 0.60 \\  \hline
11& SVM & R+DAC                     & 0.65       &0.66      &0.68  &  0.68&   0.63 &0.63 \\ \hline
12 & SVM &LCP + DAC + R        &  0.72      & 0.73    &0.69&  0.68  &  0.61 & 0.61 \\  \hline
13 & Bi LSTM  &     DAC              &  0.67    &0.68      &0.69 &0.65    &0.65& 0.66\\  \hline
14& Bi LSTM  &      ST                 &  0.66    &0.66       & 0.61   &0.67    &0.64& 0.52\\  \hline
15 & Bi LSTM  &      LCP              &  0.70    & 0.68       & 0.52 &0.52     &0.65& {\bf 0.67}\\  \hline
16 & Bi LSTM  &      R                  &  0.70    & 0.70        &0.59&0.63      &0.65&  0.66\\  \hline
17 & Bi-LSTM & LCP+ DAC        &   0.70     &  0.71      & 0.69 & 0.68     &0.61& 0.62  \\  \hline
18 & Bi-LSTM & R+ DAC             &   0.70    & 0.68      &  0.63&0.62        & 0.60& 0.64  \\  \hline
19 & Bi-LSTM & R+ LCP               &    0.69  &  0.68   &{\bf 0.71} & 0.67     & 0.64  &  0.66       \\  \hline
20 & Bi-LSTM & LCP+R +DAC    &  0.73      & {\bf 0.74}  & 0.70  &0.69 &      0.62 & 0.63\\  \hline

\end{tabular}
\caption{\label{results-tab} Results for classification on test set for each topic. Best performing model in {\bf bold.} }
\end{small}
\end{table*}
\captionsetup{font=scriptsize}

\subsection{Results}
We use standard classification evaluation measures based on
Precision/Recall and F measure. Performance evaluation uses weighted
average F-score on test set.  We first evaluate simple models based on
a single feature. 

\noindent{\bf Simple Ablation Models.}  Table~\ref{results-tab}, Rows
1A, 1B and 1C show the results for our three baseline systems. The
LexRank summarizer performs best across all topics, but overall the
results show that summarizers aimed at newswire or monologic data do
not work on argumentative dialog.

Row 3 shows that Word2Vec improves over the baseline, but this did not
work as well as it did in previous research
\cite{Habernalemnlp2015}. One reason could be that averaged Word2Vec
embeddings for each word lose too much information in long sentences.
Row 2 shows that Dialog Act Classification works better than the
random baseline for gun control and gay marriage but not for
abortion. Interestingly, Row 6 shows that
sentiment by itself beats LexRank across all topics,
suggesting a relationship of sentiment to argument
that could be further explored. 

Each Row has an additional column for each topic indicating what happens
when we first run Stanford Coreference to 
replacing each pronoun with its most representative mention.
The results show that coreference improves the
F-score for both gun control and abortion. 

LIWC
categories and Readability perform well across topics.
 
 \noindent{\bf Feature Combination Models.}\\
 We first evaluate SVM with different feature combinations, with details on results in Table~\ref{results-tab}. For the gun control topic, LIWC categories on the current sentence give an F-score of 0.72. Adding LIWC from the previous sentence improves it to 0.73 (rows 5 and 9, without coref column). In contrast, just doing a coref replacement improves LIWC current sentence score to 0.74 (row 5 for gun control, with and without coref columns). A paired t-test on the result vectors shows that coref replacement provides a statistically significant improvement at ({\bf  p \textless 0.04}).
For the Abortion topic, the overall performance is low as compared  to the other two topics suggesting that arguments used for abortion are harder to identify. Both DAC, Word2vec  scores are quite low but readability and LIWC do better.\\
The LSTM models on their own do not perform  better than  SVM across topics, but adding features to the LSTM models improves them beyond the SVM  results. We paired only LSTM (row 8) separately with the best performing model in bold for each topic in Table~\ref{results-tab} to evaluate if the combination is significant. Paired t-tests on the result vectors show that the differences in F-score are statistically significant when we compare LSTM to LSTM with features for each topic ({\bf p \textless  0.01}) for all topics, indicating that adding contextual features makes a significant improvement. 
 Adding LIWC categories from current and previous utterances to LSTM also improves performance for gun control and abortion. For the gay marriage topic, LSTM combined with LIWC and readability works better than LSTM alone. 
\subsection{Analysis and Discussion}
\label{sec:analysis}
To qualitatively gain some insight into the limitations of some of the
systems, we examined random predictions from different models.  One
reason that a Graph-based system such as LexRank performs well on DUC
might ne that DUC data sets are clustered into
related documents by human assessors. To observe the behavior of the
method on noisy data, the authors of LexRank added random documents to
each cluster to show that LexRank is
insensitive to some limited noise in the data. 
However, topic changes are more frequent in dialog and
dialogs contain content that is not necessarily related to
the argumentative purpose of the dialog. 

For example, lexical overlap is important to LexRank, but this
resulted in LexRank selecting the two of these sentences {\it Well it's not
  going to work.} and {\it Get to work!}.

One reason that  SVM with sentiment features
performs well is that positive sentiment 
predicts the not-important class. It seems that sentiment
analyzers classify both phatic
communication and sarcastic arguments as positive, both
of which can be correctly assigned to the not-important
class, as shown by the following examples:

\begin{small}
\begin{itemize}
\item  I 'll be nice ... Out of context
  sermon.
\item You 're a fine one to talk about   sliming folks
\item Yes it does
\item Sounds right to you?
\end{itemize}
\end{small}

The results show that LIWC performs well and that LIWC used to represent
context performs even better. To understand which LIWC features
were important, we
performed chi-square feature selection over LIWC features on the
training set. Content categories were highly ranked across topics,
suggesting that the LIWC features are being exploited for a form of
within-topic topic detection; this suggests that more general topic
modeling could help results.

Table~\ref{chi-sq} shows the top 5 LIWC categories for each topic
based on chi-square based feature selection on the training set for
all the three topics. Unsurprisingly, across all topics, the LIWC
marker of complexity (Words Per Sentence) appears. In addition, many
other topics link commonsense with important facets of these debates
-- the opposition in abortion between questions of the sanctity of
life (biological processes), health of individuals involved. Similarly, with Gay Marriage, we see sides
of the debate between personal relationships (family, affiliation) and
questions of sexual practice (sexual, drives). The case of Gun Control
is somewhat surprising, since one might expect to see LIWC categories
relating to life and safety. Instead we see Money category coming from discussions about gun buy back and gun prices.
To understand better why coreference resolution was helping, we also
examined cases where coreference matters. Coreference resolution can
also interact with different features such as LIWC, i.e.  since LIWC
calculates a frequency distribution of categories in the text,
corefence moves a word from the pronoun to some other category.  For
example, replacing {\it it} by {\it Government} decreases Impersonal
Pronouns and Total Pronouns, while increasing Six Letter Words. In
several cases these replacements produce correct predictions, e.g.
with \\ {\it Only if it is legal to sell it. }

% 417
\begin{table}[th!]
\begin{small}
\begin{tabular}{|p{2.0cm}|p{4.5cm}|} \hline
%\captionsetup{font=scriptsize}
Topic& LIWC Categories \\  \hline
Abortion & {\it \small  Biological Processes, Health, Second Person, Sexual, Words Per Sentence, }\\  \hline
Gun Control & {\it \small First Person Singular, Money, Second Person, Third Person Plural, Words Per Sentence }\\  \hline
Gay Marriage& {\it \small  Family, Sexual, Words Per Sentence, Affiliation, Drives }\\  \hline
\end{tabular}
%\end{scriptsize}
\end{small}
\caption{\label{chi-sq} Top 5 LIWC categories by chi-square for each topic}
\end{table}

\section{Related Work}
\label{rel-sec}
This work builds on multiple strands of research into dialog,
summarization and argumentation.

\noindent{\bf Dialog Summarization.} To the best of our knowledge,
none of the previous approaches have focused on debate dialog
summarization. Prior research on spoken dialog summarization has
explored lexical features, and information specific to meetings such
as action items, speaker status, and structural discourse
features. \cite{Zechner01,Murrayetal06,Whittakeretal12,Janinetal04,Carletta07}. In
contrast to information content, \newcite{Roman06} examine how social
phenomena such as politeness level affect summarization. Emotional
information has also been observed in summaries of professional chats
discussing technology \cite{ZhouHovy05}.  Other approaches use
semantic similarity metrics to identify the most central or important
utterances of a spoken dialog using Switchboard corpus
\cite{GurevychStrube04}.  Dialog structure and prosodic features have
been studied for finding patterns of importance and opinion
summarization on Switchboard conversations
\cite{Wang2011,ward2013patterns}. Additional parallel work is on
summarizing email thread conversations using conversational features
and dialog acts specific to the email domain
\cite{Murray08summarizingspoken,carenini2014}.

%JoshiRose07,ZhouHovy05,mieskes2007improving}
%GurevychStrube04 operated on 20 randomly chosen swbd dialogs, used semantic similarity, gold standard was 2/3 annotators said ``include'', kappa was .43 Zechner also on Switchboard. Check other stuff, also there are other papers on meeting summarization.

\noindent{\bf Summarization.} Document summarization is a mature area
of NLP, and hence spans a vast range of approaches. The graph and
clustering based systems compute sentence importance based on inter
and intra-document sentence similarities
\cite{Mihalcea04TextRank,erkan2004lexrank,Ganesan}. \cite{Carbonell1998} use
a greedy approach based on Maximal Marginal
Relevance. \cite{McDonald2007} reformulated this as a dynamic
programming problem providing a knapsack based solution. The
submodular approach by \cite{Lin2011} produces a summary by maximizing
an objective function that includes coverage and diversity.

Recently there has been a surge in data-driven approaches to summarization based on neural networks and continuous sentence features. 
An encoder decoder architecture is the main framework used in these types of models. However, one major bottleneck to applying neural network
models to extractive summarization is that the  generation systems need a huge amount of training data i.e.,
 documents with sentences labeled as summary-worthy. \cite{Nallapati2016AbstractiveTS,RushCW15,AbigailAcl2017} used models trained on the annotated version of
the Gigaword corpus and paired the first sentence of each article with its headline to form sentence-summary pairs. Such newswire models did not work well here; the neural summarization model from OpenNMT framework \cite{2017opennmt} very often generated \textless UNK \textgreater tokens for our data.
\cite{srinivas16} train an end to end neural attention model using LSTMs to summarize  source code from online programming websites. Pairing the post  title with the source code snippet from accepted answers gives a large amount of training data that can be used to generate summaries.

Our approach is similar in spirit to \cite{Li2016SIGDIAL}. In this work, RST elementary discourse units (EDU's) are used as SCU's for extractive summarization of news articles. However, we observed in debate dialogs, that the same argumentative text can be used by interlocutors on opposite sides of an issue, and  hence could not be considered in isolation as a summary unit. \newcite{Barker2016SIGDIAL} describe a corpus of original Guardian articles along with associated content (comments, groups, summaries and backlinks). However, the comment data is different from conversational dialogic debates (it is less strongly threaded, less directly dialogic, and less argumentative) and they do not present a computational model for argument summary generation. \newcite{Misraetal15} use pyramid annotation of dialog summaries on online debates to derive SCUs and labels, but they go on to work with the {\bf human-generated labels} of the pyramid annotation. Our task, using raw sentences from social media dialogs, is appreciably harder.

\noindent{\bf Argumentation.} Argumentative dialog is a highly
challenging task with creative, analytical and practical abilities
needed to persuade or convince another person, but what constitutes a
"good argument" is still an open ended question
\cite{jackson1980structure,toulmin03,Sternberg2008,Walton2008-WALAS}. The
real world arguments found in social media dialog are informal,
unstructured and so the well established argument theories may not be
a good predictor of people's choice of arguments
\cite{Habernal2014,Rosenfeld2016}.  In this work, we propose pyramid
based summarization to rank and select arguments in social media
dialog, which to the best of our knowledge is a novel method for
ranking arguments in conversational data.
% Providing Arguments in Discussions Based on the Prediction of Human Argumentative Behavior Ariel Rosenfeld and Sarit Kraus
 
\section{Conclusion and Future Work}
\label{conc-sec}
We presented a novel method for argument  summarization of dialog exchanges from social media debates with our results significantly beating the traditional summarization baselines. We show that adding context based features improves argument summarization. Since we could find both topic specific and topic independent features, we plan to explore  unsupervised topic modeling that could be used to create a larger and more diverse dataset  and build sequential models that could generalize well  across a vast range of topics.
\section*{Acknowledgments}
This work was supported by NSF CISE RI 1302668. Thanks to the three anonymous reviewers for helpful comments.
\bibliography{nl.bib,amita2017.bib}
\bibliographystyle{eacl2017}

%\appendix

%\section{Supplemental Material}
%\label{sec:supplemental}

%Appendices ({\em i.e.} supplementary material in the form of proofs, tables,or pseudo-code) 
\end{document}